\pdfoutput=1
\documentclass[11pt]{article}

\usepackage[final]{acl}

\usepackage{times}
\usepackage{latexsym}

\usepackage[T1]{fontenc}
\usepackage[utf8]{inputenc}

\usepackage{microtype}

\usepackage{inconsolata}

\usepackage{graphicx}
\usepackage{subcaption}

\usepackage[T1]{fontenc} % Better for English
\usepackage[utf8]{inputenc}
\usepackage{amsmath, amssymb}
\usepackage{caption}
\usepackage{geometry}
\usepackage{hyperref} % For clickable links (optional here)
\usepackage{cleveref}
\usepackage{xcolor} % For any custom colors (optional here)
\usepackage{booktabs} % For professional tables (leaderboard mentioned)
\usepackage{enumitem} % For better list control

\crefname{section}{Section}{Sections}
\Crefname{section}{Section}{Sections}
\crefname{appendix}{Appendix}{Appendices}
\Crefname{appendix}{Appendix}{Appendices}
\crefname{figure}{Figure}{Figures}
\Crefname{figure}{Figure}{Figures}
\crefname{table}{Table}{Tables}
\Crefname{table}{Table}{Tables}
\crefname{algorithm}{Algorithm}{Algorithms}
\Crefname{algorithm}{Algorithm}{Algorithms}

\title{DOoM: Difficult Olympiads of Math}
\author{
Ilya Kuleshov \\
HSE University  \\
\texttt{kul757.48@mail.ru} 
\And 
Ilin Pavel \\
ITMO \\ 
\texttt{ilinpashtet@yandex.ru} 
\And
Nikolay Kompanets \\ 
Vikhr Models \\
\texttt{lakomoor@gmail.com} 
\AND
Ksenia Sycheva \\
Vikhr Models \\
\texttt{KS9270022@gmail.com} 
\And
Aleksandr Nikolich \\
Vikhr Models \\
\texttt{Alexdragannikolich@gmail.com} 
}
\date{\today}

\begin{document}

\maketitle

\begin{abstract}
This paper introduces DOoM, a new open-source benchmark designed to assess the capabilities of language models in solving mathematics and physics problems in Russian. The benchmark includes problems of varying difficulty, ranging from school-level tasks to university Olympiad and entrance exam questions. In this paper we discuss the motivation behind its creation, describe dataset's structure and evaluation methodology, and present initial results from testing various models. Analysis of the results shows a correlation between model performance and the number of tokens used, and highlights differences in performance between mathematics and physics tasks.
\end{abstract}

\section{Introduction}
% The language model development community is actively interested in the reasoning capabilities of models and their ability to solve complex problems. Previously, our team released the Shlepa and ruArenahard benchmarks to evaluate Russian language models in terms of cultural knowledge and Instruction Following (IF). These tools have found widespread use among model providers.

% However, despite the availability of English-language benchmarks for mathematics, such as GSM8k \cite{cobbe2021trainingverifierssolvemath}, there are virtually no open, modern counterparts for the Russian language. GSM8k, developed by OpenAI, consists of grade school math word problems that require cognitive abilities but not necessarily complex mathematical ideas. For instance, typical GSM8k problems might involve multi-step arithmetic, finding the greatest common divisor (GCD) or least common multiple (LCM), or solving simple linear equations. An example problem could be: "Natalia sold clips to 4 of her friends. She sold 8 clips to each friend. How many clips did she sell in total?"

% Our initial idea to create a Russian benchmark based on the Demidovich problem book encountered difficulties in implementing accurate automated answer verification, especially for complex cases, which led to abandoning this concept. As a result, "DOoM" was developed—a new benchmark incorporating mathematics and physics problems collected from various sources.

With the rise of language models' reasoning abilities, interest in using them to solve complex problems has increased within the development community. This motivated development of various benchmarks designed to assess models' abilities to perform multi-step reasoning in specific domains, like math, physics, or coding. Despite the availability of English-language benchmarks for mathematics \citep{cobbe2021trainingverifierssolvemath, chernyshev2025umathuniversitylevelbenchmarkevaluating, wu2024conceptmathbilingualconceptwisebenchmark, gao2024omnimathuniversalolympiadlevel} and physics \citep{xu2025ugphysicscomprehensivebenchmarkundergraduate, feng2025physicsbenchmarkingfoundationmodels, shen2025phyxdoesmodelwits}, there are virtually no open, modern counterparts for the Russian language.

In this work, we present a new Russian-language benchmark, \textbf{DOoM} (Difficult olympiads of math), which incorporates mathematics and physics problems collected from a variety of sources. The tasks in DOoM span a wide range of difficulty levels, starting from high school exam problems and extending to university-level and olympiad-style tasks. We begin by providing an overview of related work and the motivation behind creating DOoM in \cref{sec:related_work}. In \cref{sec:benchmark}, we describe DOoM and its collection process in detail.

% plan: 
% interest in complex tasks 
% existing solutions on English, Shlepa, ruArenaHard
% their limitations (including being on english only) 
% what was done 

\section{Related Work} \label{sec:related_work}
\subsection{Math and Physics Benchmarks}
Many benchmarks have been designed to assess models' abilities to perform multi-step reasoning. MMLU \citep{hendrycks2021measuringmassivemultitasklanguage} tests models' knowledge with multiple-choice questions across 57 diverse subjects, including abstract algebra, high-school level mathematics, and physics, making it a strong general-purpose benchmark for evaluating factual recall and reasoning across domains. ARC \citep{clark2018thinksolvedquestionanswering} on the other hand, presents tasks that require more abstract and commonsense reasoning. It consists of grade school science questions that often involve multi-step inference and the application of basic knowledge to novel scenarios. In contrast to MMLU and ARC, many works have focused specifically on mathematical and physical reasoning, aiming to capture domain-specific multi-step problem-solving abilities in greater depth. \\

\noindent \textbf{Math}. One of the most popular math benchmarks is GSM8k \citep{cobbe2021trainingverifierssolvemath}, which consists of 8.5K school-level math word problems. These problems typically require performing basic arithmetic operations, finding the greatest common divisor, or solving systems of linear equations. Solving such problems requires models to demonstrate multi-step reasoning and numerical understanding. However, GSM8k does not evaluate a model’s grasp of more advanced mathematical concepts. Several works attempted to cover this gap. \citet{gao2024omnimathuniversalolympiadlevel} collected a benchmark that focuses on mathematical olympiad-level tasks exclusively, crawling problems and solutions from sources such as the IMO, IMC, and AoPS online forum, which often includes detailed human-written solutions. \citet{wu2024conceptmathbilingualconceptwisebenchmark} introduced a bilingual dataset with math problems aimed at evaluating models’ understanding of a wide range of mathematical concepts. Their benchmark is organized hierarchically, enabling fine-grained evaluation at the level of individual concepts. Recently \citet{chernyshev2025umathuniversitylevelbenchmarkevaluating} published a benchmark consisting of 1100 university-level math problems. All these works demonstrate that models achieving high scores on GSM8k struggle with these more complex tasks, indicating a significant gap in current models’ mathematical reasoning capabilities when it comes to challenging, expert-level problems. \\ 

\noindent \textbf{Physics}. Models have been evaluated in the physics domain less extensively than in mathematics so far. \citet{xu2025ugphysicscomprehensivebenchmarkundergraduate} designed a benchmark to evaluate models on undergraduate-level physics problems.  \citet{zhang2025physreasoncomprehensivebenchmarkphysicsbased} collected tasks that involve, on average, eight steps of reasoning per solution and require multi-modal comprehension, such as interpreting diagrams, graphs, and symbolic equations. \citet{shen2025phyxdoesmodelwits} also focus on multi-modal reasoning, organizing their benchmark around six core areas of physics to assess models’ ability to integrate visual and textual information in complex problem-solving settings. State-of-the-art models such as GPT-4o \citep{openai2024gpt4ocard}, GPT-4o-mini \citep{gpto4mini}, and Claude 3.7 Sonnet \citep{claude37} still struggle with these physics benchmarks, highlighting the difficulty of multi-step, multi-modal reasoning in this domain.

\subsection{Russian Reasoning Benchmarks}
Reasoning benchmarks have also emerged for the Russian language specifically. One of the first works in this direction is RussianSuperGLUE \citep{shavrina2020russiansuperglue}, which consists of nine datasets covering various natural language understanding tasks. Some works have focused on more specific properties. ruArenaHard is an adaptation of Arena-Hard \citep{li2024crowdsourced} to Russian, evaluating models' instruction-following abilities. Another benchmark, Shlepa, is designed to assess knowledge of Russian cultural context, including laws, popular music, books, and films. Both benchmarks have found widespread use among model providers.

\citet{nefedov2020datasetforevaluationofmathematical} translated the synthetically generated DeepMind Mathematics Dataset into Russian, while \citet{khrulev2025checkmatcheckinghandwrittenmathematical} created an evaluation suite for assessing models on handwritten solutions to Russian-language math problems. \citet{zhang2025physreasoncomprehensivebenchmarkphysicsbased} also included approximately 20K physics problems in Russian in their benchmark. However, overall evaluation of multi-step reasoning capabilities of language models in Russian—particularly in mathematics and physics—remains even more limited than in English. In this work, we aim to help bridge this gap.

\section{DOoM Benchmark} \label{sec:benchmark}
The DOoM benchmark consists of two datasets covering distinct knowledge domains - mathematics (62.1\%) and physics (37.9\%) - which we refer to as \textbf{RussianMath} and \textbf{RussianPhysics}, respectively. The problems span a wide range of topics and difficulty levels, from high school exams to university coursework and olympiad-level challenges. We begin by describing the data collection process in \cref{sec:data_collection}, followed by detailed descriptions of the mathematics and physics datasets in \cref{sec:math} and \cref{sec:physics}, respectively.

\subsection{Data Collection} \label{sec:data_collection}
One challenge in collecting tasks is the need for verifiable solutions. Initially, we attempted to parse the problem book by Demidovich \citep{demidovich1970problems}, which contains a large number of exercises in mathematical analysis. However, this idea was ultimately abandoned due to the lack of verifiable solutions, making \citet{demidovich1970problems} and similar books an impractical source for benchmark construction. Instead, we relied on Russian school textbooks and archives of school and university-level olympiads, which often include either official solutions or community-verified answers. \cref{tab:DOoM-math-sources,tab:DOoM-physics-sources} list all data sources and the percentage of tasks from each of them included in DOoM.

\subsection{Mathematics Section} \label{sec:math}
\begin{figure*}[ht]
    \centering
    \includegraphics[width=\textwidth]{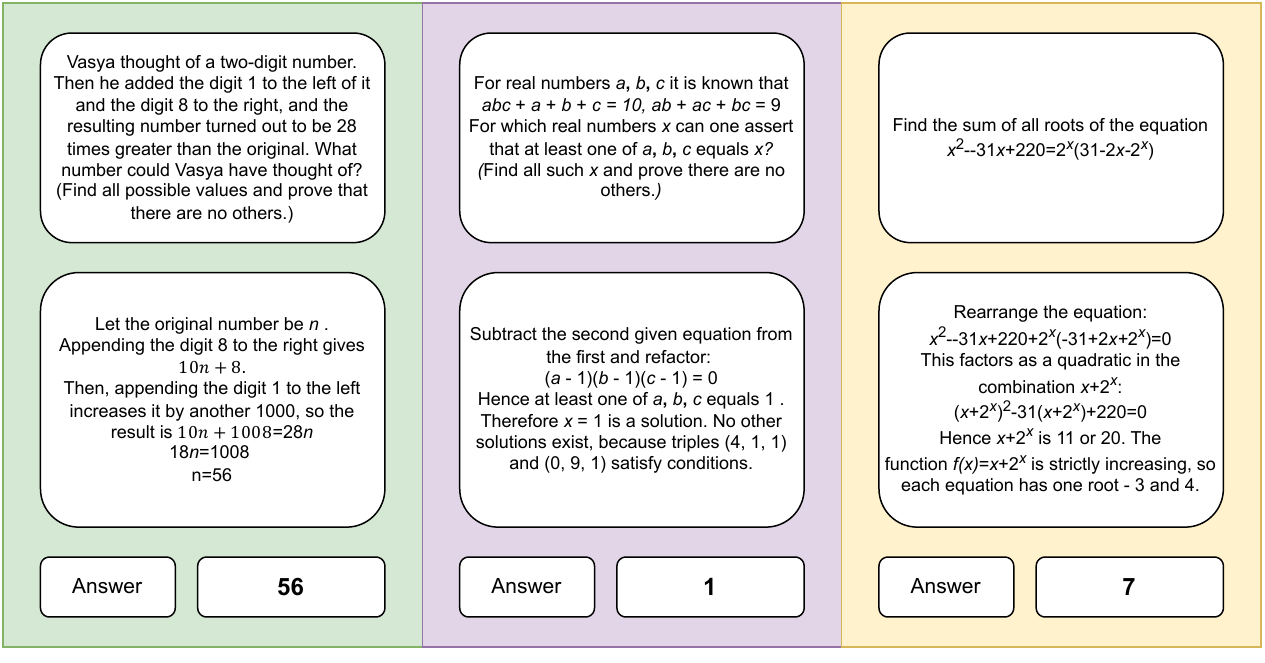}
    \caption{Examples of mathematics problems from the DOoM dataset, arranged in order of increasing difficulty from left to right: left — Grade 8 school problem; middle — HSE Olympiad, Grade 10; right — MSU Olympiad, Grade 11. All examples are translated from Russian.}
    \label{fig:math_examples}
\end{figure*}

The RussianMath dataset spans a wide range of mathematical topics, including combinatorics, algebraic and geometric progressions, and equations involving complex numbers. More advanced examples, drawn from entrance exams to leading universities, may require proving properties of intersecting parabolas given their focal parameters and positions of their focal points —tasks that demand a deep understanding of conic sections and coordinate geometry. Examples translated from Russian into English are shown in \cref{fig:math_examples}.
% The mathematics section includes problems ranging from the level of the "second part of the Unified State Exam / 11th grade" to "first-year university" and Olympiad-level problems. Example tasks might cover:
% \begin{itemize}
%     \item Combinatorics: e.g., "In how many ways can a committee of 3 be chosen from a group of 10 people?"
%     \item Geometric/Algebraic Progressions: e.g., "Find the sum of the first 10 terms of an arithmetic progression if the first term is 2 and the common difference is 3."
%     \item Equations with complex numbers: e.g., "Solve the equation $z^2 + 2z + 5 = 0$ for $z$ in the complex plane."
% \end{itemize}
% A more complex example, characteristic of supplementary entrance exams to leading universities, might involve proving properties of intersecting parabolas given their focal parameters and the positions of their foci, requiring a deep understanding of conic sections and coordinate geometry.

\begin{table}[ht]
\centering
\resizebox{0.49\textwidth}{!}{
\begin{tabular}{l c}
    \toprule
    \textbf{Math Source} & \textbf{\% of DOoM} \\
    \midrule
    School Level Olympiad & 47.7\% \\
    Regional Olympiads & 6.5\% \\ 
    MSU Entrance Problems & 4.5\% \\
    MIPT Olympiads & 10.1\% \\
    HSE Olympiads & 10.1\% \\
    Joint Interuniversity Olympiad & 19.1\% \\
    All-Russian Olympiad & 2.0\% \\
    \bottomrule
\end{tabular}
}
\caption{Math sources in DOoM. Nearly half of the math problems are sourced from school textbooks and state exam archives. The rest are drawn from regional olympiads and olympiads for school students hosted by major Russian universities.}
\label{tab:DOoM-math-sources}
\end{table}

\begin{table}[ht]
\centering
\begin{tabular}{l c}
    \toprule
    \textbf{Physics Source} & \textbf{\% of DOoM} \\
    \midrule
    School Level Olympiad & 88.2\% \\
    Regional Olympiads & 0.9\% \\ 
    1st Stage Olympiads & 0.9\% \\
    All-Russian Olympiad & 10.0\% \\
    \bottomrule
\end{tabular}
\caption{Physics sources in DOoM. The majority of physics problems are taken from school curricula and state exams. More advanced questions come from various stages of the Russian Physics Olympiad.}
\label{tab:DOoM-physics-sources}
\end{table}

% Based on the provided OCR (page 5, second pie chart), the distribution of mathematics tasks by source is approximately: School: 47.7\%, MSU: 4.5\%, Region: 6.5\%, MIPT: 10.1\%, HSE: 10.1\%, OMMO: 19.1\%, Allrus: 2.0\%.

\subsection{Physics Section} \label{sec:physics}
The RussianPhysics primarily consists of problems from various stages of the All-Russian Olympiad for school children and covers various physics domains, including mechanics, kinematics, and thermodynamics. 
% The development of this section focused on the "solvability" of tasks over a period of six months to a year, with the benchmark expected to be "saturated" within one and a half years.
% Based on the provided OCR (page 6), the distribution of physics tasks by source is approximately: School: 88.2\%, Allrus: 10.0\%, Region: 0.9\%, 1st round (presumably another Olympiad stage): 0.9\%.

% \subsection{Task Distribution}
% Overall, the benchmark (as per OCR page 5, first pie chart) is composed of approximately 62.1\% mathematics tasks and 37.9\% physics tasks.

\section{Evaluation Results}
Initial testing of several models on the DOoM benchmark has revealed interesting trends. We start by describing evaluation methodology and then discuss our findings.

\subsection{Evaluation Methodology}
\textbf{Answer Processing}. For each task, the model-generated answer is compared against a known reference solution. Since answers can be expressed in different formats, we first attempt to compare them as fractions, then as natural numbers, and finally - if both checks fail - by evaluating parsed LaTeX/Python expressions for equivalence. \\ 

\noindent \textbf{Scoring}. After processing, each answer is assigned a binary score: $1$ if it matches the reference, and $0$ otherwise. The RussianMath and RussianPhysics scores are computed as the average of binary scores across all tasks in their respective datasets. The final score is the mean of the math and physics scores.

% Model evaluation is conducted by sequentially processing tasks from the RussianMath (mathematics) and RussianPhysics (physics) datasets.

% \begin{enumerate}[label=\arabic*.]
%     \item \textbf{Answer Comparison}: For each task, the model-generated answer is compared against a known reference solution. A specialized comparison mechanism is used, which considers formatting peculiarities and precision characteristic of mathematical and physical answers.
%     \item \textbf{Binary Scoring}: Based on the comparison, each task is assigned a binary score: 1 for a correct answer and 0 for an incorrect one.
%     \item \textbf{Individual Dataset Scores}:
%     \begin{itemize}
%         \item \textbf{RussianMath Score}: Calculated as the arithmetic mean of all scores (0 or 1) obtained by the model for tasks from the mathematics dataset.
%         \item \textbf{RussianPhysics Score}: Similarly, represents the arithmetic mean of scores for all tasks from the physics dataset.
%     \end{itemize}
%     \item \textbf{Combined Score}: For models successfully tested on both datasets, an additional metric—Combined Score—is introduced. It is calculated as the simple arithmetic mean of the RussianMath Score and RussianPhysics Score.
%     \item \textbf{Data Collection}: All results (dataset scores, combined score, model name, evaluation time, number of tokens used) are systematically saved in a structured JSON file. Based on this data, a summary leaderboard is automatically generated in Markdown format.
% \end{enumerate}

\begin{table*}[ht]
\centering
\caption{Performance of different models on Russian Math and Physics tasks. 
We report scores for each subject separately as well as the overall average, and provide the total number of tokens generated by each model (including reasoning tokens). 
Within each model family, the best scores are highlighted in \textbf{bold}. 
The results reveal two notable patterns: (1) all evaluated models perform better on math than on physics, and (2) models that generate more tokens tend to achieve higher scores.}
\label{tab:model_scores}
\begin{tabular}{lcccc} 
\toprule
Model Name & Math Score & Physics Score & Overall Score & \# Tokens \\
\midrule
Gemini 2.0 Flash & 0.558 & 0.469 & 0.514 & 495,313 \\
Gemini 2.5 Pro & \textbf{0.874} & \textbf{0.582} & \textbf{0.728} & \textbf{2,227,721} \\
\midrule
OpenAI GPT-4o & 0.432 & 0.245 & 0.338 & 399,483 \\
OpenAI GPT-4.1 & 0.584 & 0.347 & 0.466 & 549,983 \\
OpenAI GPT-OSS 20B & 0.789 & 0.377 & 0.583 & 1,034,077 \\
OpenAI GPT-5-mini & 0.849 & 0.418 & 0.634 & 993,326 \\
OpenAI GPT-5-nano & 0.839 & 0.459 & 0.649 & \textbf{2,218,450} \\
OpenAI o4-mini (high reasoning) & 0.868 & 0.459 & 0.664 & 1,997,548 \\
OpenAI o3 (base) & 0.868 & 0.469 & 0.669 & 1,164,000 \\
OpenAI GPT-OSS 120B & 0.849 & \textbf{0.500} & 0.675 & 671,703 \\
OpenAI o3-mini (high reasoning) & 0.884 & \textbf{0.500} & 0.692 & 2,186,756 \\
OpenAI GPT-5 & \textbf{0.910} & \textbf{0.500} & \textbf{0.705} & 1,374,085 \\
\midrule
Claude 3.5 Sonnet & 0.416 & 0.337 & 0.376 & 252,843 \\
Claude 3.7 Sonnet & 0.542 & 0.398 & 0.470 & 405,583 \\
Claude Sonnet 4 & \textbf{0.633} & \textbf{0.469} & \textbf{0.551} & \textbf{490,996} \\
\midrule
Gemma 3 4B & 0.258 & 0.102 & 0.180 & \textbf{726,285} \\
Gemma 3 27B & \textbf{0.474} & \textbf{0.327} & \textbf{0.400} & 384,164 \\
\midrule 
Qwen2.5 72B Instruct & 0.379 & 0.000 & 0.189 & 322,441 \\
Qwen3 8B & 0.538 & 0.296 & 0.417 & 1,576,445 \\
Qwen QwQ 32B & \textbf{0.653} & \textbf{0.408} & \textbf{0.530} & \textbf{2,112,951} \\
\midrule 
DeepSeek V3-0324 & 0.432 & \textbf{0.255} & 0.343 & 339,857 \\
DeepSeek-R1 Distill (Qwen-14B) & \textbf{0.447} & 0.245 & \textbf{0.346} &  \textbf{806,258} \\
\midrule 
GigaChat 2 & 0.095 & 0.071 & 0.083 & 136,051 \\
GigaChat Max & 0.189 & 0.173 & 0.181 & 200,271 \\
GigaChat 2 Pro & 0.316 & 0.224 & 0.270 & \textbf{215,297} \\
GigaChat 2 Max & \textbf{0.363} & \textbf{0.265} & \textbf{0.314} & 185,204 \\
\bottomrule
\end{tabular}
\end{table*}

\begin{figure*}[ht]
    \centering
    % First figure using minipage
    \begin{minipage}[t]{0.48\textwidth}
        \centering
        \includegraphics[width=\linewidth]{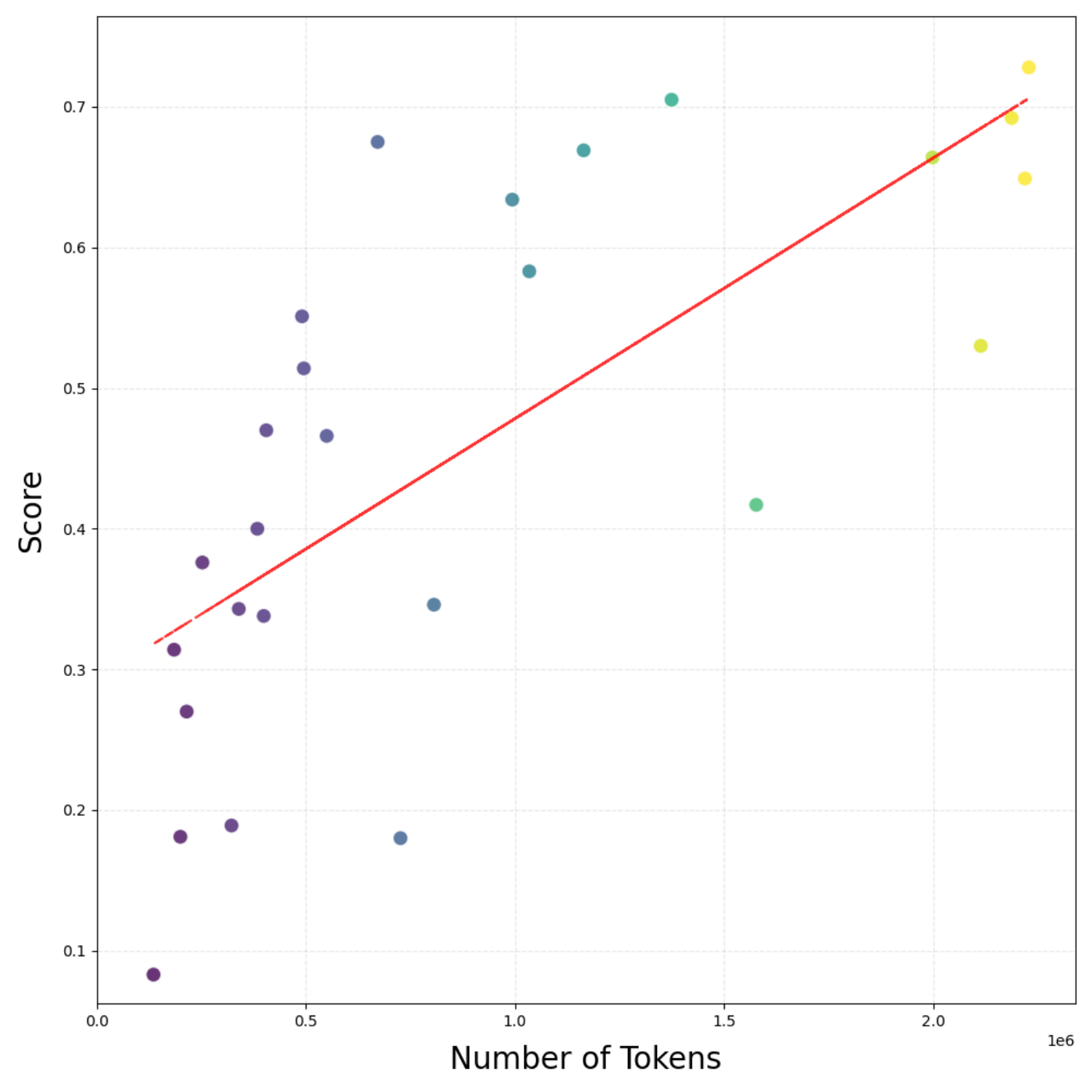}
        \captionof{figure}{Correlation between number of generated tokens and benchmark performance. The strong positive relationship (Pearson's $r = 0.707$, $p < 0.001$) indicates that more extensive reasoning is a critical factor for success.}
        \label{fig:tokens_vs_scores}
    \end{minipage}
    \hfill
    % Second figure using minipage  
    \begin{minipage}[t]{0.48\textwidth}
        \centering
        \includegraphics[width=\linewidth]{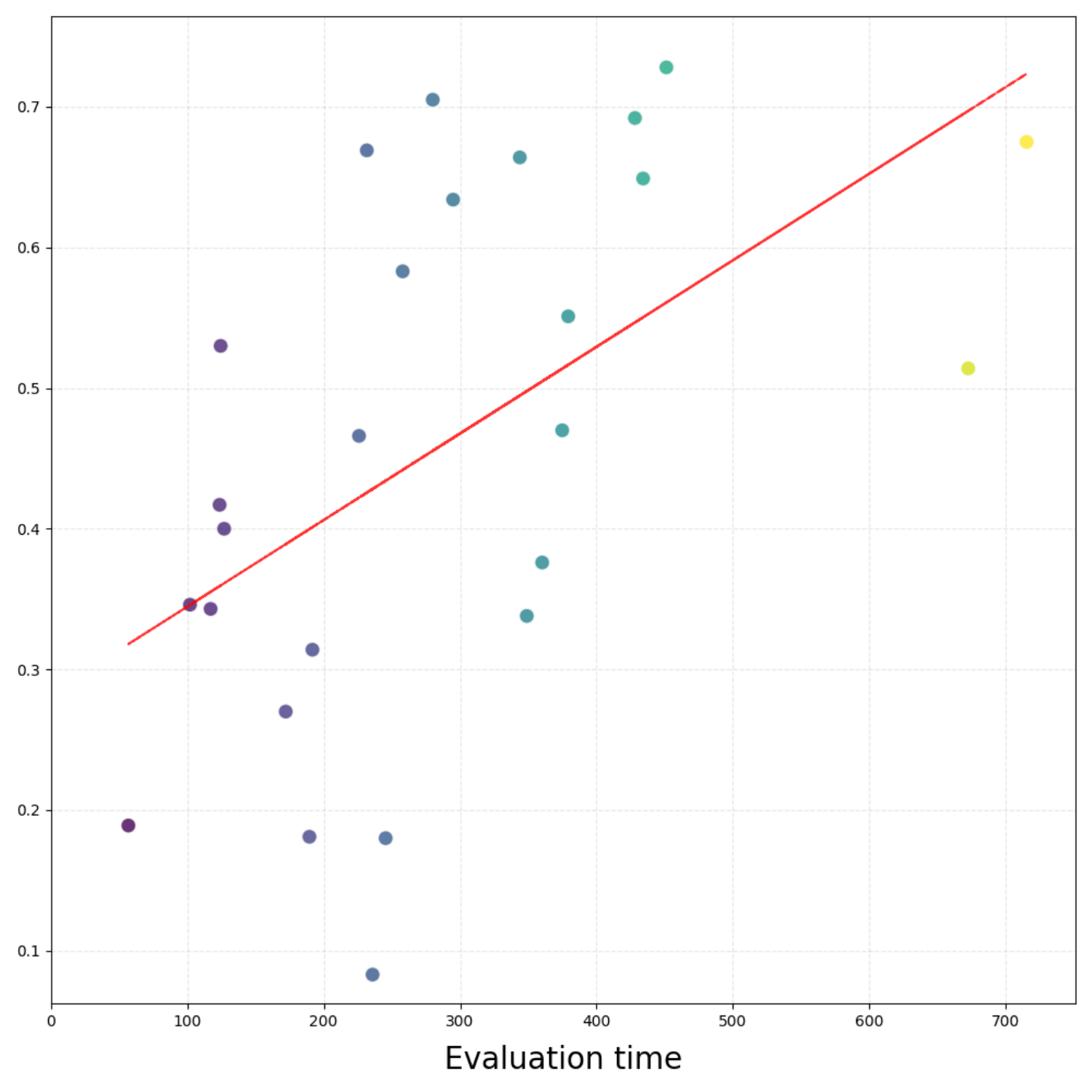}
        \captionof{figure}{Correlation between inference speed (tokens/second) and performance. The moderate positive relationship ($r = 0.531$) suggests a trade-off between reasoning thoroughness and computational efficiency.}
        \label{fig:times_vs_scores}
    \end{minipage}
\end{figure*}

% \begin{figure*}[ht]
%     \centering
%     % First subfigure
%     \begin{subfigure}[t]{0.48\textwidth}
%         \centering
%         \includegraphics[width=\textwidth]{images/tokens_vs_score.pdf}
%         % \vspace{-100pt}
%         \caption{Correlation between number of generated tokens and benchmark performance. The strong positive relationship (Pearson's $r = 0.73$, $p < 0.001$) indicates that more extensive reasoning is a critical factor for success.}
%         \label{fig:tokens_vs_scores}
%     \end{subfigure}
%     \hfill % Adds horizontal space between the figures
%     % Second subfigure
%     \begin{subfigure}[t]{0.48\textwidth}
%         \centering
%         \includegraphics[width=\textwidth]{images/times_vs_score.pdf} % Replace with your second image path
%         \caption{Correlation between generation time (measured in tokens per second) and benchmark performance. Correlation coefficient is 0.496.}
%         \label{fig:times_vs_scores}
%     \end{subfigure}
%     % \label{fig:combined_figures}
% \end{figure*}

\subsection{Overall Model Performance}
% Comparative performance of models in mathematics and physics (as suggested by OCR page 10, first graph) generally shows that most models achieve higher results on mathematics tasks compared to physics tasks. For example, a hypothetical 'Model A' might score 0.85 in Mathematics and 0.35 in Physics, while 'Model B' scores 0.55 in Mathematics and 0.28 in Physics.

To evaluate the overall capabilities of modern models in mathematical and physical reasoning, we assessed several families of API-based \citep{openai2024gpt4ocard, gpto3mini, gpto4mini, claude35sonnet} and open-source models \citep{qwen2025qwen25technicalreport, yang2025qwen3technicalreport} on the DOoM benchmark. The results, reported in \cref{tab:model_scores}, expose a significant performance gap, with all models outperforming in mathematics over physics. The highest overall score was achieved by Gemini 2.5 Pro, with 0.874 in math and 0.582 in physics. We hypothesize that this performance gap can be attributed to two primary factors: (1) the inherent complexity of physics problems, which require models to construct a qualitative mental model of a physical situation, apply relevant common-sense knowledge (e.g., friction, forces, or energy), and then select the appropriate mathematical formalism; and (2) the nature of large language models' training data, which is scraped from internet-scale sources that are inherently biased towards formal mathematics (e.g., code, proofs, equations) available in structured formats, thus providing less robust training for physics reasoning.

\subsection{Analysis of Token Usage}
% A correlation is observed between the number of tokens used by a model and its performance.
% The relationship between results (correct answer rate) and the number of tokens (OCR page 7, first graph) shows a positive correlation for both mathematics (correlation coefficient r=0.79, p=0.000) and physics (r=0.60, p=0.008).

% The overall score also demonstrates a strong positive correlation with the number of tokens used (r=0.74, p=0.000), as suggested by OCR page 7, second graph. Models employing a larger number of tokens (on the order of 2 million) generally achieve higher results. Token usage varies across models, with some models using significantly more tokens than others (e.g., 'Model X' using 405.8K tokens, 'Model Y' 222.2K tokens, as per OCR page 10, second graph).

% The relationship between token usage and model performance can also be assessed through an efficiency metric (score per 1000 tokens), as indicated by OCR page 8, first graph. This reveals that while higher token usage often leads to better scores, the efficiency (score gained per token) might vary.

In addition to reporting scores, we also measured the total number of tokens generated during evaluation (\cref{tab:model_scores}). As shown in \cref{fig:tokens_vs_scores}, there is a clear positive correlation between the number of tokens and performance, with correlation coefficients of 0.79 for math and 0.60 for physics. When allowed to generate freely without limiting the number of tokens, models producing more tokens (around two million) consistently achieve higher scores. Notably, the number of tokens varies substantially across models, ranging from 100,000 to 2,000,000.

\subsection{Analysis of Processing Speed}
% The relationship between processing speed (tokens per second) and model performance (OCR page 8, second graph) highlights the trade-offs between inference speed and the quality of solutions. Some models may process tokens faster but achieve lower scores, while others might be slower but more accurate.

Unlike the clear relationship with token count, the connection between processing speed and performance is less straightforward. We observe a moderate positive correlation ($r = 0.50$); however, this trend is not universal. The results reveal notable outliers, including high-accuracy models with comparatively slow processing speeds and vice versa, indicating that raw inference speed alone is not a reliable predictor of overall capability.

\section{Conclusion}
This paper has introduced the DOoM benchmark, a comprehensive open-source tool for assessing the capabilities of language models in solving mathematics and physics problems in Russian. By providing a structured dataset of problems ranging from school-level to Olympiad difficulty, DOoM addresses a pronounced lack of resources for evaluating reasoning skills in the Russian language.

Our initial evaluation of various models reveals two key insights: a strong correlation between performance and the volume of tokens generated during reasoning, and a significant performance gap between mathematics and physics tasks, suggesting the latter presents a uniquely complex challenge.

We offer DOoM to the global AI research community as a foundation for future work. We anticipate expanding the benchmark's scope and complexity and welcome collaborative efforts to ensure it remains a robust and evolving standard for evaluating reasoning capabilities.

\bibliography{main}

@misc{cobbe2021trainingverifierssolvemath,
      title={Training Verifiers to Solve Math Word Problems}, 
      author={Karl Cobbe and Vineet Kosaraju and Mohammad Bavarian and Mark Chen and Heewoo Jun and Lukasz Kaiser and Matthias Plappert and Jerry Tworek and Jacob Hilton and Reiichiro Nakano and Christopher Hesse and John Schulman},
      year={2021},
      eprint={2110.14168},
      archivePrefix={arXiv},
      primaryClass={cs.LG},
      url={https://arxiv.org/abs/2110.14168}, 
}

@misc{shen2025phyxdoesmodelwits,
      title={PhyX: Does Your Model Have the "Wits" for Physical Reasoning?}, 
      author={Hui Shen and Taiqiang Wu and Qi Han and Yunta Hsieh and Jizhou Wang and Yuyue Zhang and Yuxin Cheng and Zijian Hao and Yuansheng Ni and Xin Wang and Zhongwei Wan and Kai Zhang and Wendong Xu and Jing Xiong and Ping Luo and Wenhu Chen and Chaofan Tao and Zhuoqing Mao and Ngai Wong},
      year={2025},
      eprint={2505.15929},
      archivePrefix={arXiv},
      primaryClass={cs.AI},
      url={https://arxiv.org/abs/2505.15929}, 
}

@misc{feng2025physicsbenchmarkingfoundationmodels,
      title={PHYSICS: Benchmarking Foundation Models on University-Level Physics Problem Solving}, 
      author={Kaiyue Feng and Yilun Zhao and Yixin Liu and Tianyu Yang and Chen Zhao and John Sous and Arman Cohan},
      year={2025},
      eprint={2503.21821},
      archivePrefix={arXiv},
      primaryClass={cs.AI},
      url={https://arxiv.org/abs/2503.21821}, 
}

@misc{zhang2025physreasoncomprehensivebenchmarkphysicsbased,
      title={PhysReason: A Comprehensive Benchmark towards Physics-Based Reasoning}, 
      author={Xinyu Zhang and Yuxuan Dong and Yanrui Wu and Jiaxing Huang and Chengyou Jia and Basura Fernando and Mike Zheng Shou and Lingling Zhang and Jun Liu},
      year={2025},
      eprint={2502.12054},
      archivePrefix={arXiv},
      primaryClass={cs.AI},
      url={https://arxiv.org/abs/2502.12054}, 
}

@misc{xu2025ugphysicscomprehensivebenchmarkundergraduate,
      title={UGPhysics: A Comprehensive Benchmark for Undergraduate Physics Reasoning with Large Language Models}, 
      author={Xin Xu and Qiyun Xu and Tong Xiao and Tianhao Chen and Yuchen Yan and Jiaxin Zhang and Shizhe Diao and Can Yang and Yang Wang},
      year={2025},
      eprint={2502.00334},
      archivePrefix={arXiv},
      primaryClass={cs.CL},
      url={https://arxiv.org/abs/2502.00334}, 
}

@misc{gao2024omnimathuniversalolympiadlevel,
      title={Omni-MATH: A Universal Olympiad Level Mathematic Benchmark For Large Language Models}, 
      author={Bofei Gao and Feifan Song and Zhe Yang and Zefan Cai and Yibo Miao and Qingxiu Dong and Lei Li and Chenghao Ma and Liang Chen and Runxin Xu and Zhengyang Tang and Benyou Wang and Daoguang Zan and Shanghaoran Quan and Ge Zhang and Lei Sha and Yichang Zhang and Xuancheng Ren and Tianyu Liu and Baobao Chang},
      year={2024},
      eprint={2410.07985},
      archivePrefix={arXiv},
      primaryClass={cs.CL},
      url={https://arxiv.org/abs/2410.07985}, 
}

@misc{wu2024conceptmathbilingualconceptwisebenchmark,
      title={ConceptMath: A Bilingual Concept-wise Benchmark for Measuring Mathematical Reasoning of Large Language Models}, 
      author={Yanan Wu and Jie Liu and Xingyuan Bu and Jiaheng Liu and Zhanhui Zhou and Yuanxing Zhang and Chenchen Zhang and Zhiqi Bai and Haibin Chen and Tiezheng Ge and Wanli Ouyang and Wenbo Su and Bo Zheng},
      year={2024},
      eprint={2402.14660},
      archivePrefix={arXiv},
      primaryClass={cs.CL},
      url={https://arxiv.org/abs/2402.14660}, 
}

@misc{khrulev2025checkmatcheckinghandwrittenmathematical,
      title={CHECK-MAT: Checking Hand-Written Mathematical Answers for the Russian Unified State Exam}, 
      author={Ruslan Khrulev},
      year={2025},
      eprint={2507.22958},
      archivePrefix={arXiv},
      primaryClass={cs.CV},
      url={https://arxiv.org/abs/2507.22958}, 
}

@inbook{nefedov2020datasetforevaluationofmathematical,
    author = {Nefedov, Mikhail},
    year = {2020},
    month = {09},
    pages = {135-144},
    title = {Dataset for Evaluation of Mathematical Reasoning Abilities in Russian},
    isbn = {978-3-030-59081-9},
    doi = {10.1007/978-3-030-59082-6_10}
}

@inproceedings{shavrina2020russiansuperglue,
   title={RussianSuperGLUE: A Russian Language Understanding Evaluation Benchmark},
   url={http://dx.doi.org/10.18653/v1/2020.emnlp-main.381},
   DOI={10.18653/v1/2020.emnlp-main.381},
   booktitle={Proceedings of the 2020 Conference on Empirical Methods in Natural Language Processing (EMNLP)},
   publisher={Association for Computational Linguistics},
   author={Shavrina, Tatiana and Fenogenova, Alena and Anton, Emelyanov and Shevelev, Denis and Artemova, Ekaterina and Malykh, Valentin and Mikhailov, Vladislav and Tikhonova, Maria and Chertok, Andrey and Evlampiev, Andrey},
   year={2020} }

@misc{hendrycks2021measuringmassivemultitasklanguage,
      title={Measuring Massive Multitask Language Understanding}, 
      author={Dan Hendrycks and Collin Burns and Steven Basart and Andy Zou and Mantas Mazeika and Dawn Song and Jacob Steinhardt},
      year={2021},
      eprint={2009.03300},
      archivePrefix={arXiv},
      primaryClass={cs.CY},
      url={https://arxiv.org/abs/2009.03300}, 
}

@misc{clark2018thinksolvedquestionanswering,
      title={Think you have Solved Question Answering? Try ARC, the AI2 Reasoning Challenge}, 
      author={Peter Clark and Isaac Cowhey and Oren Etzioni and Tushar Khot and Ashish Sabharwal and Carissa Schoenick and Oyvind Tafjord},
      year={2018},
      eprint={1803.05457},
      archivePrefix={arXiv},
      primaryClass={cs.AI},
      url={https://arxiv.org/abs/1803.05457}, 
}

@misc{chernyshev2025umathuniversitylevelbenchmarkevaluating,
      title={U-MATH: A University-Level Benchmark for Evaluating Mathematical Skills in LLMs}, 
      author={Konstantin Chernyshev and Vitaliy Polshkov and Ekaterina Artemova and Alex Myasnikov and Vlad Stepanov and Alexei Miasnikov and Sergei Tilga},
      year={2025},
      eprint={2412.03205},
      archivePrefix={arXiv},
      primaryClass={cs.CL},
      url={https://arxiv.org/abs/2412.03205}, 
}

@misc{gpto3mini,
  author = {OpenAI},
  title = {OpenAI o3-mini: Pushing the frontier of cost-effective reasoning},
  howpublished = {\url{https://openai.com/index/openai-o3-mini/}},
  year = {2025}
}

@misc{gpto4mini,
  author = {OpenAI},
  title = {Introducing OpenAI o3 and o4-mini},
  howpublished = {\url{https://openai.com/index/introducing-o3-and-o4-mini/}},
  year = {2025}
}

@misc{claude37,
  author = {Anthropic},
  title = {Claude 3.7 Sonnet and Claude Code},
  howpublished = {\url{https://www.anthropic.com/news/claude-3-7-sonnet}},
  year = {2025}
}

@misc{claude35sonnet,
  author = {Anthropic},
  title = {Introducing Claude 3.5 Sonnet},
  howpublished = {\url{https://www.anthropic.com/news/claude-3-5-sonnet}},
  year = {2024}
}

@misc{openai2024gpt4ocard,
      title={GPT-4o System Card}, 
      author={OpenAI},
      year={2024},
      eprint={2410.21276},
      archivePrefix={arXiv},
      primaryClass={cs.CL},
      url={https://arxiv.org/abs/2410.21276}, 
}

@article{li2024crowdsourced,
  title={From Crowdsourced Data to High-Quality Benchmarks: Arena-Hard and BenchBuilder Pipeline},
  author={Li, Tianle and Chiang, Wei-Lin and Frick, Evan and Dunlap, Lisa and Wu, Tianhao and Zhu, Banghua and Gonzalez, Joseph E and Stoica, Ion},
  journal={arXiv preprint arXiv:2406.11939},
  year={2024}
}

@book{demidovich1970problems,
  author    = {B. P. Demidovich},
  title     = {Problems in Mathematical Analysis},
  year      = {1970},
  publisher = {Mir Titles},
  address   = {Moscow},
  note      = {Translated from Russian},
}

@misc{qwen2025qwen25technicalreport,
      title={Qwen2.5 Technical Report}, 
      author={An Yang and Baosong Yang and Beichen Zhang and Binyuan Hui and Bo Zheng and Bowen Yu and Chengyuan Li and Dayiheng Liu and Fei Huang and Haoran Wei and Huan Lin and Jian Yang and Jianhong Tu and Jianwei Zhang and Jianxin Yang and Jiaxi Yang and Jingren Zhou and Junyang Lin and Kai Dang and Keming Lu and Keqin Bao and Kexin Yang and Le Yu and Mei Li and Mingfeng Xue and Pei Zhang and Qin Zhu and Rui Men and Runji Lin and Tianhao Li and Tianyi Tang and Tingyu Xia and Xingzhang Ren and Xuancheng Ren and Yang Fan and Yang Su and Yichang Zhang and Yu Wan and Yuqiong Liu and Zeyu Cui and Zhenru Zhang and Zihan Qiu},
      year={2025},
      eprint={2412.15115},
      archivePrefix={arXiv},
      primaryClass={cs.CL},
      url={https://arxiv.org/abs/2412.15115}, 
}

@misc{yang2025qwen3technicalreport,
      title={Qwen3 Technical Report}, 
      author={An Yang and Anfeng Li and Baosong Yang and Beichen Zhang and Binyuan Hui and Bo Zheng and Bowen Yu and Chang Gao and Chengen Huang and Chenxu Lv and Chujie Zheng and Dayiheng Liu and Fan Zhou and Fei Huang and Feng Hu and Hao Ge and Haoran Wei and Huan Lin and Jialong Tang and Jian Yang and Jianhong Tu and Jianwei Zhang and Jianxin Yang and Jiaxi Yang and Jing Zhou and Jingren Zhou and Junyang Lin and Kai Dang and Keqin Bao and Kexin Yang and Le Yu and Lianghao Deng and Mei Li and Mingfeng Xue and Mingze Li and Pei Zhang and Peng Wang and Qin Zhu and Rui Men and Ruize Gao and Shixuan Liu and Shuang Luo and Tianhao Li and Tianyi Tang and Wenbiao Yin and Xingzhang Ren and Xinyu Wang and Xinyu Zhang and Xuancheng Ren and Yang Fan and Yang Su and Yichang Zhang and Yinger Zhang and Yu Wan and Yuqiong Liu and Zekun Wang and Zeyu Cui and Zhenru Zhang and Zhipeng Zhou and Zihan Qiu},
      year={2025},
      eprint={2505.09388},
      archivePrefix={arXiv},
      primaryClass={cs.CL},
      url={https://arxiv.org/abs/2505.09388}, 
}

\end{document}